\documentclass[sigconf]{acmart}

\AtBeginDocument{%
  }

\acmConference[NLPIR '22]{6th International Conference on Natural Language Processing and Information Retrieval}{December 16-18,  2022}{Bangkok, Thailand}

\usepackage[nolist]{acronym}
\usepackage{graphicx}
\usepackage{multirow}
\usepackage{bm}
\usepackage{algorithm}
\usepackage[noend]{algpseudocode}
\usepackage{stfloats}
\usepackage{float}
\usepackage{multicol}

\setcopyright{none}
\renewcommand\footnotetextcopyrightpermission[1]{} 
\begin{document}

\title[Evaluating Unsupervised Text Classification: Zero-shot and Similarity-based Approaches]{Evaluating Unsupervised Text Classification: \\ Zero-shot and Similarity-based Approaches}

\author{Tim Schopf}
\orcid{0000-0003-3849-0394}
\affiliation{%
  \institution{Technical University of Munich, Department of Computer Science}
  \city{Garching}
  \country{Germany}}
\email{tim.schopf@tum.de}

\author{Daniel Braun}
\orcid{0000-0001-8120-3368}
\affiliation{%
  \institution{University of Twente, Department of High-tech Business and Entrepreneurship}
  \city{Enschede}
  \country{Netherlands}}
\email{d.braun@utwente.nl}

\author{Florian Matthes}
\orcid{0000-0002-6667-5452}
\affiliation{%
  \institution{Technical University of Munich, Department of Computer Science}
  \city{Garching}
  \country{Germany}}
\email{matthes@tum.de}

\renewcommand{\shortauthors}{Schopf et al.}

\begin{abstract}
    Text classification of unseen classes is a challenging Natural Language Processing task and is mainly attempted using two different types of approaches. Similarity-based approaches attempt to classify instances based on similarities between text document representations and class description representations. Zero-shot text classification approaches aim to generalize knowledge gained from a training task by assigning appropriate labels of unknown classes to text documents. Although existing studies have already investigated individual approaches to these categories, the experiments in literature do not provide a consistent comparison. This paper addresses this gap by conducting a systematic evaluation of different similarity-based and zero-shot approaches for text classification of unseen classes. Different state-of-the-art approaches are benchmarked on four text classification datasets, including a new dataset from the medical domain. Additionally, novel \mbox{SimCSE} \cite{gao-etal-2021-simcse} and SBERT-based \cite{reimers-gurevych-2019-sentence} baselines are proposed, as other baselines used in existing work yield weak classification results and are easily outperformed. Finally, the novel similarity-based Lbl2TransformerVec approach is presented, which outperforms previous state-of-the-art approaches in unsupervised text classification. Our experiments show that similarity-based approaches significantly outperform zero-shot approaches in most cases. Additionally, using SimCSE or SBERT embeddings instead of simpler text representations increases similarity-based classification results even further.
\end{abstract}



\maketitle

\begin{acronym}
\acro{nlp}[NLP]{Natural Language Processing}
\acro{ke}[KE]{keyword enrichment}
\acro{plm}[PLM]{Pretrained Language Model}
\acroplural{plm}[PLMs]{Pretrained Language Models}
\acro{zsl}[ZSL]{zero-shot learning}
\acro{zstc}[0SHOT-TC]{zero-shot text classification}
\acro{esa}[ESA]{Explicit Semantic Analysis}
\acro{lsa}[LSA]{Latent Semantic Analysis}
\acro{lsi}[LSI]{Latent Semantic Indexing}
\acro{svd}[SVD]{Singular Value Decomposition}
\end{acronym}

\section{Introduction}

Unsupervised text classification approaches aim to perform categorization without using annotated data during training and therefore offer the potential to reduce annotation costs. Despite this possibility, unsupervised text classification approaches have attracted significantly less attention in contrast to supervised text classification approaches. As a result, extensive work is already being done to structure and evaluate the field of text classification with a focus on supervised approaches \cite{MIRONCZUK201836, Kadhim2019, info10040150, 10.1145/3439726} while little research has been conducted on evaluating unsupervised text classification approaches. This study bridges this gap by assessing the two most popular categories of unsupervised text classification approaches.

Generally, unsupervised text classification approaches aim to map text to labels based on their textual description, without using annotated training data. To accomplish this, there exist mainly two categories of approaches. The first category can be summarized under \textit{similarity-based} approaches. Thereby, the approaches generate semantic embeddings of both the texts and the label descriptions, before attempting to match the texts to the labels using similarity measures such as cosine similarity \cite{Song_Roth_2014, zero-shot-semsim, haj-yahia-etal-2019-towards, webist21}. The second category uses \textit{\ac{zsl}} to classify texts of unseen classes. \ac{zsl} uses labeled training instances belonging to seen classes to learn a classifier that can predict testing instances belonging to different, unseen classes \cite{10.1145/3293318}. Although \ac{zsl} techniques employ annotated data for training, they do not use labels to provide information about the target classes and can use their knowledge of the previously seen classes to classify instances of unseen classes. Since pretrained \acf{zstc} models do not require training or fine-tuning on labeled data from the target classes, we classify them as an unsupervised text classification strategy. The highly successful deep learning performances of recent years have also stimulated research initiatives for \ac{zstc} \cite{Pushp2017TrainOT, rios-kavuluru-2018-shot, zhang-etal-2019-integrating, yin-etal-2019-benchmarking, 9411914}. We argue, that one of the main differences between \ac{zsl} and similarity-based approaches is, that \ac{zsl} approaches use annotated data for seen classes to predict texts of unseen classes, whereas pure similarity-based approaches do not require seen classes at all.

We summarize the contributions of our work as follows::
\begin{itemize}
    \item We evaluate the \textit{similarity-based} and \textit{\acl{zsl}} categories for unsupervised text classification of topics. Thereby, we conduct experiments with representative approaches of each category on four different benchmark datasets, including a new text classification dataset from the medical domain.
    \item We propose simple but strong baselines for unsupervised text classification based on SimCSE \cite{gao-etal-2021-simcse} and SBERT \cite{reimers-gurevych-2019-sentence} embedding similarities. Previous work has mostly been evaluated against different weak baselines such as Word2Vec \cite{NIPS2013_9aa42b31} similarities which are easy to outperform and tend to overestimate the performance of new unsupervised text classification approaches.
    \item Since transformer-based text representations have been widely established as state-of-the-art for semantic text similarity in recent years, we further adapt Lbl2Vec \cite{webist21, schopf_lbl2vec_23}, one of the most recent and well-performing similarity-based methods for unsupervised text classification, to be used with transformer-based language models\footnote{{Code available: \url{https://github.com/sebischair/Lbl2Vec}}}.
\end{itemize}

\section{Related Work}

\citet{Chang2008ImportanceOS} investigated unsupervised text classification under the umbrella name "Dataless Classification" in one of their earliest works. They used \ac{esa} \cite{10.5555/1625275.1625535} to embed the text and label descriptions in a common semantic space and picked the label with the highest matching score for classification. Semantic embeddings are vector representations of texts that capture their semantic meaning and can be used as input for a variety of different \ac{nlp} downstream tasks \cite{10.1145/3460824.3460826,schneider-etal-2022-decade,schopf_etal_kdir22,10.1145/3535782.3535835}. Dataless classification is based on the idea that semantic representations of labels are equally relevant as learning semantic text representations and was subsequently further examined in \cite{Song_Roth_2014, 10.5555/2886521.2886630, li-etal-2016-joint, 10.5555/3060832.3061027}.

With the progress of text embeddings, the term "Dataless Classification" became less prevalent and was rather represented by the broad category of similarity-based approaches for unsupervised classification. Within this category, \citet{zero-shot-semsim} embedded text documents and textual label descriptions with Word2Vec and used cosine similarity between text and label embeddings to predict instances of unseen classes. \citet{haj-yahia-etal-2019-towards} proposed to enrich label descriptions with expert keywords and subsequently conduct unsupervised classification based on \ac{lsa} \cite{Deerwester1990IndexingBL} similarities. \citet{Stammbach2021DocSCANUT} introduced DocSCAN, which produces semantic representations of text documents and uses Semantic Clustering by Adopting Nearest-Neighbors for unsupervised text classification. \citet{webist21} used Doc2Vec \cite{pmlr-v32-le14} to jointly embed word, document, and label vectors for subsequent similarity-based unsupervised text classification. 

Similarly, \citet{AAAI1612058} jointly embedded document, label, and word representations with Doc2Vec. However, they learned a ranking function for multi-label classification and attempted to predict instances of unseen classes in a zero-shot setting for classification. \citet{zhang-etal-2019-integrating} integrated four types of semantic knowledge (word embeddings, class descriptions, class hierarchy, and a general knowledge graph) in a two-phase framework for \ac{zstc}. \citet{yin-etal-2019-benchmarking} proposed to treat \ac{zstc} as a textual entailment problem, while \citet{ye-etal-2020-zero} tackled \ac{zstc} with a semi-supervised self-training approach.
 
\section{Text Classification Approaches}

\subsection{Baselines}\label{sec:baselines}

We compare the findings of current state-of-the-art unsupervised text classification approaches to some basic baselines to evaluate their performance. \\ \\
\textbf{\ac{lsa}:} \ac{svd} is used on term-document matrices to learn a set of concepts (or topics) related to the documents and terms \cite{Deerwester1990IndexingBL}. For each dataset, we apply \ac{lsa} to learn $n=$ \textit{number of classes} concepts. Afterward, the text documents are classified according to the highest cosine similarity of resulting \ac{lsa} vectors of documents and label keywords. A similar approach was used by \citet{haj-yahia-etal-2019-towards} for unsupervised text classification.\\ \\
\textbf{Word2Vec:} This produces semantic vector representations of words based on surrounding context words \cite{NIPS2013_9aa42b31}. A Skip-gram model with a vector size of 300 and a surrounding window of 5 is trained for each dataset. The average of word embeddings is then used to represent the text documents and label keywords. The text documents are predicted according to the highest cosine similarity of the resulting Word2Vec representations of documents and label keywords for classification. Similar approaches were used by \citet{yin-etal-2019-benchmarking} and \citet{ye-etal-2020-zero} as baseline for \ac{zstc}.\\ \\
\textbf{SimCSE:} This is a contrastive learning framework that produces sentence embeddings which acieve state-of-the-art results in semantic similarity tasks \cite{gao-etal-2021-simcse}. Algorithm \ref{alg:split_docs} is first used to separate the text documents into paragraphs because SimCSE models have a maximum input sequence length. Then, the average of SimCSE paragraph embeddings as text document representations and the average of SimCSE label keyword embeddings as class representations are employed. Finally, the text documents are classified according to the highest cosine similarity of the resulting SimCSE representations of document and label keywords. \\ \\ 
\textbf{SBERT:} This is a modification of BERT \cite{devlin-etal-2019-bert} that uses siamese and triplet network structures to derive semantically meaningful sentence embeddings \cite{reimers-gurevych-2019-sentence}. We use the same classification approach as described in the paragraph above, except that we now use SBERT embeddings instead of SimCSE embeddings. 

\begin{algorithm}
    \caption{Split text document into paragraphs}
    \label{alg:split_docs}
    \begin{algorithmic}[0]
    \Require
    \State {$d=$ text document}
    \State {$m_k=$ max input sequence length of transformer-model $k$}
    \State {\textit{len($x$)} = numer of words in text $x$}
    \Procedure{split-document}{$d, m_k$} 
        \State {$\textrm{sentences}_d \gets$ sentence\_tokenize($d$)}
        \State {$\textrm{paragraphs}_d \gets \emptyset$}
        \State {$p \gets \emptyset$}
        \For{$s$ \textbf{in} sentences}
           \If{\textit{len($p$)} + \textit{len($s$)} < $\frac{m_k}{2}$} 
                \State {$p \gets p$ + $s$}
            \Else
                \State {$\textrm{paragraphs}_d \gets \textrm{paragraphs}_d$ + $p$}
                \State {$p \gets \emptyset$}
            \EndIf
        \EndFor
        \Return $\textrm{paragraphs}_d$
    \EndProcedure
    \end{algorithmic}
\end{algorithm}

\subsection{Similarity-based Text Classification}\label{sec:similarity-based-classsification}

As previously stated, numerous similarity-based approaches for unsupervised text classification exist. However, the recently introduced Lbl2Vec approach \cite{webist21} is focused on in this study. We chose Lbl2Vec to represent the similarity-based classification category since preliminary experiments confirmed improved performance compared with other similarity-based approaches. Lbl2Vec works by jointly embedding word, document, and label representations. First, word and documented representations are learned with Doc2Vec. Then, the average of label keyword representations for each class is used to find a set of most similar candidate document representations via cosine similarity. The average of candidate document representations, in turn, generates the label vector for each class. For classification, eventually, the documents are assigned to the class where the cosine similarity of the label vector and the document vector is the highest.

We adapt the Lbl2Vec approach, using transformer-based text representations instead of Doc2Vec to create jointly embedded word, document, and label representations. Since transformer-based text representations currently achieve state-of-the-art results in text-similarity tasks, we investigate the effect of the different resulting text representations on this similarity-based text classification strategy. In this paper, we use SimCSE \cite{gao-etal-2021-simcse} and SBERT \cite{reimers-gurevych-2019-sentence} transformer-models to create text representations. We use the average paragraph embeddings per document as document representations. The paragraphs of documents are obtained by applying Algorithm \ref{alg:split_docs}. To find candidate documents for label vectors, the transformer-models create individual embeddings for each label keyword. Then, cosine similarity is used to find the documents that are most similar to the average of the label keyword embeddings for each class. After obtaining the candidate documents this way, the label vectors as an average of the candidate document representations for each class are computed. For classification, the documents are assigned to the class where the cosine similarity between the label vector and the document vector is the highest. In the following, the Lbl2Vec approach adapted with transformer-based text representations is referred to as Lbl2TransformerVec.

\subsection{Zero-shot Text Classification}\label{sec:zero-shot-entailment}

\ac{zstc} is still relatively less researched, but nevertheless yields some promising approaches. Using pretrained \ac{zstc} models can be considered an unsupervised text classification strategy, since no label information of target classes are required for training or fine-tuning. Although newer approaches such as the one of \citet{9411914} exist, preliminary experiments confirmed that the zero-shot entailment approach \cite{yin-etal-2019-benchmarking} still produces state-of-the-art \ac{zstc} results in predicting instances of unseen classes. As the name already implies, the zero-shot entailment approach deals with \ac{zstc} as a textual entailment problem. The underlying idea is similar to that of similarity-based text classification approaches. Conventional \ac{zstc} classifiers fail to understand the actual problem since the label names are usually converted into simple indices. Therefore, these classifiers can hardly generalize from seen to unseen classes. Considering \ac{zstc} as an entailment problem, however, provides the classifier with a textual label description and therefore enables it to understand the meaning of labels. 

Similarly, TARS \cite{halder-etal-2020-task} also uses the textual label description to classify text in a zero-shot setting. However, TARS approaches the task as a binary classification problem, where a text and a textual label description is given to the model, which makes a prediction about whether that label is true or not. The TARS authors state that this approach significantly outperforms GPT-2 \cite{Radford2019LanguageMA} in \ac{zstc}.

Since the zero-shot entailment approach currently produces state-of-the-art results in predicting instances of unseen classes and TARS also promises encouraging results, we select both approaches to represent the \ac{zsl} category for unsupervised text classification.

\section{Datasets}\label{sec:datasets}

Our evaluation is based on four text classification datasets from different domains. As we use the semantic meaning of class descriptions for unsupervised text classification, we infer label keywords from each class name that serves the purpose of textual class descriptions. Thereby, the inference step simply consists of using the class names provided by the official documentation of the datasets as label keywords. In a few cases, we additionally substituted the class names with synonymous or semantically similar keywords, if we considered this to be a more appropriate description of a certain class.

\subsection{20Newsgroups}\label{sec:20news}
The 20Newsgroups\footnote{\href{http://qwone.com/~jason/20Newsgroups}{qwone.com/~jason/20Newsgroups}} dataset is a common text classification benchmark dataset. It was introduced by \citet{Lang95} and comprised approximately 20,000 newsgroup posts, equally distributed across 20 different newsgroups classes. Appendix \ref{sec:appendix_20news} summarizes the classes and inferred label keywords.

\subsection{AG's Corpus}\label{sec:agcorpus}
The original AG's Corpus\footnote{\href{http://groups.di.unipi.it/~gulli/AG_corpus_of_news_articles}{groups.di.unipi.it/{\raise.17ex\hbox{$\scriptstyle\sim$}}gulli/AG\_corpus\_of\_news\_articles}} dataset is a collection of over 1 million news articles on different topics. The \citet{10.5555/2969239.2969312} version is used in this study, which was constructed by choosing the 4 largest classes from the original corpus. Each class contains 30,000 training samples and 1,900 testing samples. In total, the dataset consists of 127,600 samples. Appendix \ref{sec:appendix_ag} summarizes the classes and inferred label keywords.

\subsection{Yahoo! Answers}\label{sec:yahoo}
The Yahoo! Answers dataset was constructed by \citet{10.5555/2969239.2969312} and contains 10 different topic classes. Each class contains 140,000 training samples and 6,000 testing samples. In total, the dataset consists of 1,460,000 samples. Appendix \ref{sec:appendix_yahoo} summarizes the classes and inferred label keywords.

\subsection{Medical Abstracts}\label{sec:medical}
We obtained the raw Medical Abstracts dataset through Kaggle\footnote{\href{https://www.kaggle.com/datasets/chaitanyakck/medical-text}{https://www.kaggle.com/datasets/chaitanyakck/medical-text}}. The original corpus contains 28.880 medical abstracts describing 5 different classes of patient conditions, with only about half of the dataset being annotated. Furthermore, the original annotations consist of numerical labels only. A medical text classification dataset from this corpus by using only the labeled medical abstracts was created, adding descriptive labels to the respective classes, and splitting the data into a training set and a test set. Table \ref{tab:medical_data} shows a summary of the processed Medical Abstracts dataset.

\begin{table}[ht]
    \centering
    \scalebox{0.9}{
    \begin{tabular}{|l||l|l|l|}
    \hline
    \multicolumn{1}{|c||}{\textbf{Class Name}} & \multicolumn{1}{c|}{\textbf{\#training}} & \multicolumn{1}{c|}{\textbf{\#test}} & \multicolumn{1}{c|}{$\sum$} \\ \hline\hline
    \multirow{2}{*}{\textbf{Neoplasms}}                        & \multirow{2}{*}{2530}                                   & \multirow{2}{*}{633}                                & \multirow{2}{*}{3163}                              \\ & & & \\ \hline
    \textbf{\begin{tabular}[c]{@{}l@{}}Digestive system\\ diseases\end{tabular}}        & 1195                                   & 299                                & 1494                              \\ \hline
    \textbf{\begin{tabular}[c]{@{}l@{}}Nervous system\\ diseases\end{tabular}}          & 1540                                   & 385                                & 1925                              \\ \hline
    \textbf{\begin{tabular}[c]{@{}l@{}}Cardiovascular\\ diseases\end{tabular}}          & 2441                                   & 610                                & 3051                              \\ \hline
    \textbf{\begin{tabular}[c]{@{}l@{}}General pathological\\ conditions\end{tabular}}  & 3844                                   & 961                                & 4805                              \\ \hline\hline
    $\sum$                                      & 11550                                  & 2888                               & 14438                             \\ \hline
    \end{tabular}}
    \caption{\label{tab:medical_data}Class distributions within the Medical Abstracts dataset.}
\end{table}

The inferred label keywords for each class are summarized in Appendix \ref{sec:appendix_medical}. We make this corpus available under the Creative Commons CC BY-SA 3.0 license\footnote{\href{https://creativecommons.org/licenses/by-sa/3.0/}{https://creativecommons.org/licenses/by-sa/3.0}} at \url{https://github.com/sebischair/Medical-Abstracts-TC-Corpus}.

\section{Experimental Design}
For evaluation of different unsupervised text classification approaches, we use the datasets described in Section \ref{sec:datasets}. Since we don't use label information to train the classifiers, we concatenate the training and test sets for each dataset and use the respective entire concatenated datasets for training and testing. After checking the Yahoo! Answers dataset for consistency, we observe that some answers we try to classify are empty or contain simple yes/no statements. Therefore, answers that are empty or consist of only one word are removed. We use the label keywords described in Appendix \ref{sec:appendix} for all text classification approaches to create class representations. Additionally, for the baselines and similarity-based approaches, we use the average of the respective label keyword embeddings as class representations. In contrast, for the zero-shot approaches, the respective label keywords of the 20Newsgroups, AG's Corpus, and Yahoo! Answers classes are concatenated with "and" and then used as hypotheses/label descriptions. For the Medical Abstracts dataset just the class names are used as hypotheses/label descriptions. 

We use the approaches described in Section \ref{sec:baselines} as baselines for unsupervised text classification. For our SimCSE experiments, we use the \textit{
sup-simcse-roberta-large}\footnote{\href{https://huggingface.co/princeton-nlp/sup-simcse-roberta-large}{princeton-nlp/sup-simcse-roberta-large}} model. To create embeddings for the SBERT baseline approach, we use two different pretrained SBERT models. We choose the general purpose models \textit{all-mpnet-base-v2}\footnote{\href{https://huggingface.co/sentence-transformers/all-mpnet-base-v2}{sentence-transformers/all-mpnet-base-v2}} and \textit{all-MiniLM-L6-v2}\footnote{\href{https://huggingface.co/sentence-transformers/all-MiniLM-L6-v2}{sentence-transformers/all-MiniLM-L6-v2}}, trained on more than one billion training pairs and expected to perform well on sentence similarity tasks. The \textit{all-mpnet-base-v2} model is larger than the  \textit{all-MiniLM-L6-v2} model and guarantees slightly better quality sentence embeddings. The smaller \textit{all-MiniLM-L6-v2} model, on the other hand, guarantees a five times faster encoding time while still providing sentence embeddings of high quality. 

For evaluation of similarity-based text classification, we apply the approaches described in Section \ref{sec:similarity-based-classsification}. Similar to the SimCSE and SBERT baseline approaches, we generate text embeddings for the Lbl2TransformerVec approach using the \textit{
sup-simcse-roberta-large}, \textit{all-mpnet-base-v2}, and \textit{all-MiniLM-L6-v2} models.

For evaluation of \ac{zstc}, we use the zero-shot approaches described in Section \ref{sec:zero-shot-entailment}. We conduct experiments with three different pretrained zero-shot entailment models: a DeBERTa \cite{DBLP:journals/corr/abs-2006-03654} model \footnote{\href{https://huggingface.co/MoritzLaurer/DeBERTa-v3-base-mnli-fever-docnli-ling-2c}{MoritzLaurer/DeBERTa-v3-base-mnli-fever-docnli-ling-2c}} trained on the MultiNLI \cite{williams-etal-2018-broad}, Fever-NLI \cite{thorne-etal-2018-fever}, LingNLI \cite{parrish-etal-2021-putting-linguist}, and DocNLI \cite{yin-etal-2021-docnli} datasets, a large BART \cite{lewis-etal-2020-bart} model\footnote{\href{https://huggingface.co/facebook/bart-large-mnli}{facebook/bart-large-mnli}} trained on the MultiNLI dataset, and a smaller DistilBERT \cite{Sanh2019DistilBERTAD} model\footnote{\href{https://huggingface.co/typeform/distilbert-base-uncased-mnli}{typeform/distilbert-base-uncased-mnli}} trained on the MultiNLI dataset. For TARS experiments, we use the BERT-based pretrained \textit{tars-base-v8}\footnote{\href{https://flair.informatik.hu-berlin.de/resources/models/tars-base/}{https://flair.informatik.hu-berlin.de/resources/models/tars-base/}} model. Since \textit{tars-base-v8} pretraining is partly done on AG's Corpus, we don't conduct TARS experiments on this dataset.

\begin{table*}[!b]
    \centering
    \newcommand{\STAB}[1]{\begin{tabular}{@{}c@{}}#1\end{tabular}}
    \begin{tabular}{ll||c|c|c|c}
     &                                                                                           & \textbf{20Newsgroups} & \textbf{AG's Corpus}  & \textbf{Yahoo! Answers} & \textbf{Medical Corpus} \\ \hline\hline
      \multirow{10}{*}{\STAB{\rotatebox[origin=c]{90}{Baselines}}} & \multirow{2}{*}{\textbf{LSA}}                                                                              & \multirow{2}{*}{17.89}                 & \multirow{2}{*}{41.17}                 & \multirow{2}{*}{15.82}                   & \multicolumn{1}{c|}{\multirow{2}{*}{31.61}}                   \\
     &                                                                                           & \multicolumn{1}{l|}{} & \multicolumn{1}{l|}{} & \multicolumn{1}{l|}{}   & \multicolumn{1}{l|}{}    \\ \cline{2-6} 
     & \multirow{2}{*}{\textbf{Word2Vec}}                                                                         & \multirow{2}{*}{12.87}                 & \multirow{2}{*}{28.22}                 & \multirow{2}{*}{12.55}                   & \multicolumn{1}{c|}{\multirow{2}{*}{25.00}}                   \\
     &                                                                                           & \multicolumn{1}{l|}{} & \multicolumn{1}{l|}{} & \multicolumn{1}{l|}{}   & \multicolumn{1}{l|}{}   \\ \cline{2-6} 
     & \multirow{2}{*}{\textbf{SimCSE}}                                                                         & \multirow{2}{*}{42.84}                 & \multirow{2}{*}{80.10}                 & \multirow{2}{*}{49.90}                   & \multicolumn{1}{c|}{\multirow{2}{*}{34.94}}                   \\
     &                                                                                           & \multicolumn{1}{l|}{} & \multicolumn{1}{l|}{} & \multicolumn{1}{l|}{}   & \multicolumn{1}{l|}{}    \\ \cline{2-6} 
     & \textbf{\begin{tabular}[c]{@{}l@{}}SBERT\\ (all-MiniLM-L6-v2)\end{tabular}}              & 57.89                 & 68.57                 & 43.77                   & \multicolumn{1}{c|}{46.53}                   \\ \cline{2-6} 
     & \textbf{\begin{tabular}[c]{@{}l@{}}SBERT\\ (all-mpnet-base-v2)\end{tabular}}             & 59.75                 & 70.84                 & 51.25                   & \multicolumn{1}{c|}{46.34}                   \\ \hline\hline
     \multirow{8}{*}{\STAB{\rotatebox[origin=c]{90}{\small{Similarity-based TC}}}} & \multirow{2}{*}{\textbf{Lbl2Vec}}                                                                          & \multirow{2}{*}{\textbf{65.71}}        & \multirow{2}{*}{74.63}                & \multirow{2}{*}{44.26}                   & \multicolumn{1}{c|}{\multirow{2}{*}{43.03}}                   \\ 
     &                                                                                           & \multicolumn{1}{l|}{} & \multicolumn{1}{l|}{} & \multicolumn{1}{l|}{}   & \multicolumn{1}{l|}{}  \\ \cline{2-6} 
     & \textbf{\begin{tabular}[c]{@{}l@{}}Lbl2TransformerVec\\ (SimCSE)\end{tabular}} & 58.79                 & \textbf{83.79}                 & 53.32          & \multicolumn{1}{c|}{39.60}     \\ \cline{2-6} 
     & \textbf{\begin{tabular}[c]{@{}l@{}}Lbl2TransformerVec\\ (all-MiniLM-L6-v2)\end{tabular}}  & 63.01                 & 80.88        & 52.87                   & \multicolumn{1}{c|}{54.57}                   \\ \cline{2-6} 
     & \textbf{\begin{tabular}[c]{@{}l@{}}Lbl2TransformerVec\\ (all-mpnet-base-v2)\end{tabular}} & 64.69                 & 80.05                 & \textbf{55.84}          & \multicolumn{1}{c|}{56.46}                               \\ \hline\hline
     \multirow{8}{*}{\STAB{\rotatebox[origin=c]{90}{\small{\ac{zstc}}}}} & \multirow{2}{*}{\textbf{TARS}}      & \multirow{2}{*}{17.65}                 & \multirow{2}{*}{-}                 & \multirow{2}{*}{34.60}                   & \multicolumn{1}{c|}{\multirow{2}{*}{10.92}}                   \\ 
     &                                                                                           & \multicolumn{1}{l|}{} & \multicolumn{1}{l|}{} & \multicolumn{1}{l|}{}   & \multicolumn{1}{l|}{}    \\ \cline{2-6}
     & \textbf{\begin{tabular}[c]{@{}l@{}}Zero-shot Entailment\\ (DistilBERT)\end{tabular}}      & 16.27                 & 59.48                 & 31.81                   & \multicolumn{1}{c|}{25.74}                   \\ \cline{2-6} 
     & \textbf{\begin{tabular}[c]{@{}l@{}}Zero-shot Entailment\\ (BART-large)\end{tabular}}      & 38.54                 & 68.24                 & 40.21                   & \multicolumn{1}{c|}{56.86}    \\ \cline{2-6} 
     & \textbf{\begin{tabular}[c]{@{}l@{}}Zero-shot Entailment\\ (DeBERTa)\end{tabular}}      & 47.19                 & 72.57                 & 43.09                   & \multicolumn{1}{c|}{\textbf{57.28}}  \\ \hline     
    \end{tabular}
    \caption{\label{tab:f1_evaluation_scores}$\text{F}_{1}\text{-scores}$ (micro) of examined text classification approaches on different datasets. The best results on the respective dataset are displayed in bold. Since we use micro-averaging to calculate our classification metrics, we realize equal $\text{F}_{1}$, Precision, and Recall scores respectively.}
\end{table*}

\subsection{Hypotheses}

We had four main hypotheses prior to conducting the experiments.

\begin{enumerate}
    \item \textbf{\ac{zstc} models yield better text classification results than similarity-based approaches:} \\
    The \ac{zstc} models investigated in this paper use a cross-encoder architecture which allows them to compare the input text and the textual label description simultaneously, while performing self-attention over both. In contrast, the similarity-based approaches encode the input text and label description separately. For semantic text similarity tasks, cross-encoders have proven to perform better than calculating cosine similarities for separately encoded texts. Hence we expect a similar outcome for unsupervised text classification.
    \item \textbf{Using larger \acp{plm} results in better classification performances:} \\
    Although this may seem obvious, we nevertheless want to examine whether the outcomes of using larger \acp{plm} justify their drawbacks during training and inference. 
    \item \textbf{Classification results of \ac{plm}-based approaches are highly domain dependent:} \\
    We assume that, \ac{plm}-based approaches lose some of their classification performance when dealing with very domain-specific corpora, since this specific domain may be underrepresented in the training data. Therefore, we anticipate that for certain domains, approaches like Lbl2Vec that trains unsupervised models on the classification data from scratch might perform comparably better. 
    \item \textbf{With increasing length of text documents, the performance of SimCSE and SBERT-based approaches decreases:} \\
    SimCSE and SBERT representations are most effective if the texts are embedded as a whole and no truncation strategy is used. Since we compute the document representations as the average of their respective paragraph embeddings, we assume that the quality of SimCSE and SBERT document embeddings decreases with increasing text length, resulting in worse classification performance accordingly. 
\end{enumerate}

\section{Evaluation}\label{sec:evaluation}

Table \ref{tab:f1_evaluation_scores} shows the performances of unsupervised text classification approaches for each dataset, measured in $\text{F}_{1}\text{-scores}$. We can observe that none of the baselines achieves the highest $\text{F}_{1}\text{-score}$ on any dataset based on these data. This indicates that the use of advanced unsupervised text classification approaches usually yields better results than simple baseline approaches. However, we observe that the \ac{lsa} and Word2Vec approaches generally yield the worst results and are easy to outperform. In contrast, the SimCSE and SBERT baselines produce strong $\text{F}_{1}\text{-scores}$ that even some of the advanced approaches could not surpass in certain cases. Furthermore, the SimCSE and SBERT baseline approaches may produce better results than the Lbl2Vec similarity-based approach on three datasets. We nevertheless can deduce that the use of advanced similarity-based approaches generally produces better unsupervised text classification results than the use of simple baseline approaches. Specifically, the Lbl2TransformerVec approaches using SBERT embeddings appear to be promising, as they consistently perform well across all datasets and outperform the baseline results. In contrast, the \ac{zstc} approaches perform consistently weak and in the majority of cases did not even manage to outperform the baseline results. However, the DeBERTa zero-shot entailment model could classify the domain-specific medical abstracts surprisingly well and achieved the best $\text{F}_{1}\text{-score}$ of all classifiers on this dataset. Nevertheless, considering that all \ac{zstc} models yielded disappointing results in all remaining experiments and also failed to outperform the baselines, our first hypothesis can be rejected.

Concerning our second hypothesis, the results are less obvious. On the one hand, the large DeBERTa zero-shot entailment model always significantly outperforms the smaller BART-large and DistilBERT zero-shot entailment models. Additionally, the BERT-based TARS model performs slightly better than the smaller DistilBERT zero-shot entailment model, except in case of the domain-specific Medical Abstracts dataset. Conversely, \textit{all-mpnet-base-v2} and \textit{all-MiniLM-L6-v2}-based approaches tend to produce unsupervised classification results that are fairly close to each other. Although these results are quite similar and sometimes even approaches based on the smaller \textit{all-MiniLM-L6-v2} model perform better, we nevertheless see that approaches based on the larger \textit{all-mpnet-base-v2} produce slightly better results in most cases. Therefore, we find sufficient support for our second hypothesis in the case of similarity-based unsupervised text classification approaches, with even stronger support in case of \ac{zstc}.

\begin{figure}[!hb]
    \centering
    \includegraphics[width=\linewidth]{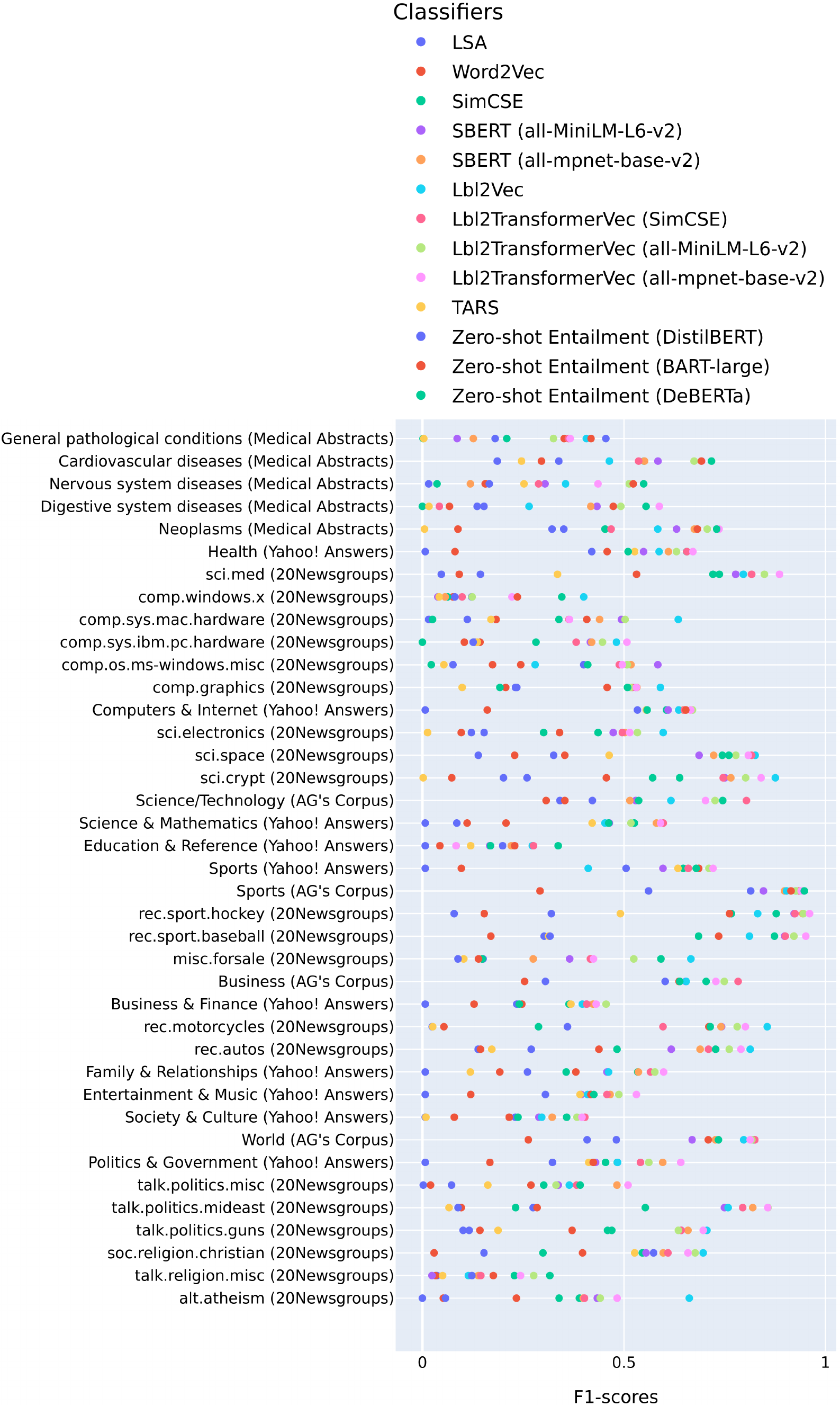}
    \caption{$\text{F}_{1}\text{-scores}$ of classification models for the individual classes of all four benchmark datasets.}
    \label{fig:class-wise_f1}
\end{figure}

\newpage

Figure \ref{fig:class-wise_f1} shows a more detailed view of the classification results by visualizing the $\text{F}_{1}\text{-scores}$ of classification models for the individual classes of all datasets. Here we observe that the overall performance of classifiers is class-dependent. While all classifiers generally yield good results for some classes (e.g. the sports classes of the 20Newsgroups and AG's Corpus datasets), all classifiers performed considerably worse for other classes (e.g. "talk.religion.misc [20Newsgroups]" or "Education \& Reference [Yahoo! Answers"]). When we compare the performance of the Lbl2Vec model, which was trained from scratch, to that of \ac{plm}-based approaches, we discover that all approaches produce similar results for many classes. In some classes, however, Lbl2Vec clearly outperforms $\text{F}_{1}\text{-scores}$ of all other \ac{plm}-based approaches (e.g. in the "comp.sys.mac.hardware", "misc.forsale", or "alt.atheism" classes of the 20Newsgroups dataset). Unfortunately, this fact can't be generalized from individual classes to  the entire domains. For example, Lbl2Vec scores relatively well in "comp.sys.mac.hardware (20Newsgroups)" and "comp.windows.x (20Newsgroups)" classes, but performs significantly worse than PLM-based models in "comp.os.ms-windows.misc (20Newsgroups)", despite all classes belonging to the same domain. We conclude that although a model trained from scratch can yield better results than \ac{plm}-based approaches in some cases, as demonstrated by the Lbl2Vec results on the 20Newsgroups dataset, we do not find sufficient support for our third hypothesis.

\begin{table}[!ht]
    \centering
    \begin{tabular}{|l|c|c|}
    \hline
    \textbf{Model}                                                                            & \multicolumn{1}{l|}{\textbf{Kendall's $\bm{\tau}$}} & \multicolumn{1}{l|}{\textbf{p-value}} \\ \hline
    \multirow{2}{*}{SimCSE}              & \multirow{2}{*}{-0.16}                                       &  \multirow{2}{*}{0.16}   \\                               
      &  &  \\ \hline
    \begin{tabular}[c]{@{}l@{}}SBERT\\ (all-MiniLM-L6-v2)\end{tabular}              & 0.07                                       & 0.52                                  \\ \hline
    \begin{tabular}[c]{@{}l@{}}SBERT\\ (all-mpnet-base-v2)\end{tabular}           & 0.04                                      & 0.73                                  \\ \hline
    \begin{tabular}[c]{@{}l@{}}Lbl2TransformerVec\\ (SimCSE)\end{tabular}  & -0.08                                      & 0.46                                  \\ \hline
    \begin{tabular}[c]{@{}l@{}}Lbl2TransformerVec\\ (all-MiniLM-L6-v2)\end{tabular}  & -0.03                                      & 0.82                                  \\ \hline
    \begin{tabular}[c]{@{}l@{}}Lbl2TransformerVec\\ (all-mpnet-base-v2)\end{tabular} & 0.03                                      & 0.80                                  \\ \hline
    \end{tabular}
    \caption{Results of the correlation analysis to measure the relationship between ${X=}$ average number of document words of each class in all four benchmark datasets and ${Y=}$ $\text{F}_{1}\text{-scores}$ of each class in all four benchmark datasets.}
    \label{tab:correlation_analysis}
\end{table}

To test our fourth hypothesis, we perform a correlation analysis measuring monotonic relationships between the $\text{F}_{1}\text{-scores}$ of the transformer-based classification approaches per class and the average number of document words per class. We choose Kendall's $\tau$ as correlation coefficient, because of its robustness against outliers and the small dataset. Further, we determine a significance level of 0.05. Table \ref{tab:correlation_analysis} shows the results of this correlation analysis. We can observe that all correlation coefficients are close to zero. Therefore, we can't identify a correlation trend. Moreover, all p-values exceed our defined significance level of 0.05 by far, indicating our test results are statistically insignificant. As a result, we find no support for our fourth hypothesis and reject it.

\section{Limitations}
One significant limitation of this evaluation is that only unsupervised text classification results for the \textit{topic} aspect are considered. This means that we consider classification results based on topics that describe what a text document is about only. However, text classification can be seen in a broader context where aspects such as \textit{emotion} or \textit{situation} are predicted as well \cite{yin-etal-2019-benchmarking}. We only focus on unsupervised similarity-based approaches and \ac{zstc} approaches that can classify the entire datasets without requiring training or fine-tuning on parts of the datasets. Self-training approaches which address the problem as a semi-supervised task or \ac{zsl} approaches that use parts of the datasets for training or fine-tuning, may lead to different results. Although we try to generalize from the datasets and approaches examined in the experiments, our evaluation is limited to those datasets and approaches nonetheless.

\section{Conclusion}
The evaluation of unsupervised text classification approaches in Section \ref{sec:evaluation}, has shown that similarity-based approaches generally outperform \ac{zstc} approaches in a variety of different domains. \ac{zstc} approaches tend to produce relatively bad results and are therefore hardly eligible for unsupervised text classification problems. In comparison, similarity-based approaches appear to predict instances of unknown classes more accurately. The characteristics of text embeddings enable representations of similar topics or classes to be located close to each other in embedding space. This implies that text representation approaches which are able to cluster topics in embedding space coherently also perform well in unsupervised text classification. This characteristic is also evident in our work. DensMap \cite{Narayan2020DensityPreservingDV} visualizations of document representations in embedding space used for classification in this work are shown in Appendix \ref{sec:appendix_umap} in Figure \ref{fig:densmap}. To improve similarity-based text classification results even further, we can use additional, different, or more descriptive label keywords than the ones we used for evaluation \cite{haj-yahia-etal-2019-towards, webist21}.

We showed that using larger \acp{plm} yield better results for 0SHOT-TC, but this is not always the case for similarity-based approaches. Therefore, unsupervised text classification using smaller \acp{plm} can be conducted in order to benefit from faster inference without necessarily sacrificing much performance in terms of $\text{F}_{1}\text{-score}$.

Our evaluation shows that simple approaches such as \ac{lsa} or Word2Vec are easy to outperform and therefore are not recommended to be used as baselines for text classification of unseen classes. However, our proposed SimCSE and SBERT baseline approaches generate strong unsupervised text classification results, outperforming even some more advanced classifiers. Therefore, we propose to use SimCSE and SBERT baselines for evaluating unsupervised text classification approaches and \ac{zstc} performance on unseen classes in future work. 

Lbl2TransformerVec, our proposed similarity-based text classification approach yields best $\text{F}_{1}\text{-scores}$ for almost all datasets. This is largely due to the great text-similarity characteristics of SimCSE and SBERT representations. Therefore, we believe that future unsupervised text classification work will benefit considerably from enhanced text embedding representations.


\bibliographystyle{ACM-Reference-Format}
\bibliography{references, anthology}


\begin{thebibliography}{45}


\ifx \showCODEN    \undefined \def \showCODEN     #1{\unskip}     \fi
\ifx \showDOI      \undefined \def \showDOI       #1{#1}\fi
\ifx \showISBNx    \undefined \def \showISBNx     #1{\unskip}     \fi
\ifx \showISBNxiii \undefined \def \showISBNxiii  #1{\unskip}     \fi
\ifx \showISSN     \undefined \def \showISSN      #1{\unskip}     \fi
\ifx \showLCCN     \undefined \def \showLCCN      #1{\unskip}     \fi
\ifx \shownote     \undefined \def \shownote      #1{#1}          \fi
\ifx \showarticletitle \undefined \def \showarticletitle #1{#1}   \fi
\ifx \showURL      \undefined \def \showURL       {\relax}        \fi
\providecommand\bibfield[2]{#2}
\providecommand\bibinfo[2]{#2}
\providecommand\natexlab[1]{#1}
\providecommand\showeprint[2][]{arXiv:#2}

\bibitem[Braun et~al\mbox{.}(2021)]%
        {10.1145/3460824.3460826}
\bibfield{author}{\bibinfo{person}{Daniel Braun}, \bibinfo{person}{Oleksandra
  Klymenko}, \bibinfo{person}{Tim Schopf}, \bibinfo{person}{Yusuf Kaan~Akan},
  {and} \bibinfo{person}{Florian Matthes}.} \bibinfo{year}{2021}\natexlab{}.
\newblock \showarticletitle{The Language of Engineering: Training a
  Domain-Specific Word Embedding Model for Engineering}. In
  \bibinfo{booktitle}{\emph{2021 3rd International Conference on Management
  Science and Industrial Engineering}} (Osaka, Japan)
  \emph{(\bibinfo{series}{MSIE 2021})}. \bibinfo{publisher}{Association for
  Computing Machinery}, \bibinfo{address}{New York, NY, USA},
  \bibinfo{pages}{8–12}.
\newblock
\showISBNx{9781450388887}
\urldef\tempurl%
\url{https://doi.org/10.1145/3460824.3460826}
\showDOI{\tempurl}


\bibitem[Chang et~al\mbox{.}(2008)]%
        {Chang2008ImportanceOS}
\bibfield{author}{\bibinfo{person}{Ming-Wei Chang}, \bibinfo{person}{Lev-Arie
  Ratinov}, \bibinfo{person}{Dan Roth}, {and} \bibinfo{person}{Vivek
  Srikumar}.} \bibinfo{year}{2008}\natexlab{}.
\newblock \showarticletitle{Importance of Semantic Representation: Dataless
  Classification}. In \bibinfo{booktitle}{\emph{AAAI}}.
  \bibinfo{pages}{830--835}.
\newblock
\urldef\tempurl%
\url{https://www.aaai.org/Library/AAAI/2008/aaai08-132.php}
\showURL{%
\tempurl}


\bibitem[Chen et~al\mbox{.}(2015)]%
        {10.5555/2886521.2886630}
\bibfield{author}{\bibinfo{person}{Xingyuan Chen}, \bibinfo{person}{Yunqing
  Xia}, \bibinfo{person}{Peng Jin}, {and} \bibinfo{person}{John Carroll}.}
  \bibinfo{year}{2015}\natexlab{}.
\newblock \showarticletitle{Dataless Text Classification with Descriptive LDA}.
  In \bibinfo{booktitle}{\emph{Proceedings of the Twenty-Ninth AAAI Conference
  on Artificial Intelligence}} (Austin, Texas)
  \emph{(\bibinfo{series}{AAAI'15})}. \bibinfo{publisher}{AAAI Press},
  \bibinfo{pages}{2224–2231}.
\newblock
\showISBNx{0262511290}
\urldef\tempurl%
\url{https://www.aaai.org/ocs/index.php/AAAI/AAAI15/paper/view/9524}
\showURL{%
\tempurl}


\bibitem[Deerwester et~al\mbox{.}(1990)]%
        {Deerwester1990IndexingBL}
\bibfield{author}{\bibinfo{person}{Scott~C. Deerwester},
  \bibinfo{person}{Susan~T. Dumais}, \bibinfo{person}{Thomas~K. Landauer},
  \bibinfo{person}{George~W. Furnas}, {and} \bibinfo{person}{Richard~A.
  Harshman}.} \bibinfo{year}{1990}\natexlab{}.
\newblock \showarticletitle{Indexing by Latent Semantic Analysis}.
\newblock \bibinfo{journal}{\emph{J. Am. Soc. Inf. Sci.}}  \bibinfo{volume}{41}
  (\bibinfo{year}{1990}), \bibinfo{pages}{391--407}.
\newblock
\urldef\tempurl%
\url{https://cis.temple.edu/~vasilis/Courses/CIS750/Papers/deerwester90indexing_9.pdf}
\showURL{%
\tempurl}


\bibitem[Devlin et~al\mbox{.}(2019)]%
        {devlin-etal-2019-bert}
\bibfield{author}{\bibinfo{person}{Jacob Devlin}, \bibinfo{person}{Ming-Wei
  Chang}, \bibinfo{person}{Kenton Lee}, {and} \bibinfo{person}{Kristina
  Toutanova}.} \bibinfo{year}{2019}\natexlab{}.
\newblock \showarticletitle{{BERT}: Pre-training of Deep Bidirectional
  Transformers for Language Understanding}. In
  \bibinfo{booktitle}{\emph{Proceedings of the 2019 Conference of the North
  {A}merican Chapter of the Association for Computational Linguistics: Human
  Language Technologies, Volume 1 (Long and Short Papers)}}.
  \bibinfo{publisher}{Association for Computational Linguistics},
  \bibinfo{address}{Minneapolis, Minnesota}, \bibinfo{pages}{4171--4186}.
\newblock
\urldef\tempurl%
\url{https://doi.org/10.18653/v1/N19-1423}
\showDOI{\tempurl}


\bibitem[Gabrilovich and Markovitch(2007)]%
        {10.5555/1625275.1625535}
\bibfield{author}{\bibinfo{person}{Evgeniy Gabrilovich} {and}
  \bibinfo{person}{Shaul Markovitch}.} \bibinfo{year}{2007}\natexlab{}.
\newblock \showarticletitle{Computing Semantic Relatedness Using
  Wikipedia-Based Explicit Semantic Analysis}. In
  \bibinfo{booktitle}{\emph{Proceedings of the 20th International Joint
  Conference on Artifical Intelligence}} (Hyderabad, India)
  \emph{(\bibinfo{series}{IJCAI'07})}. \bibinfo{publisher}{Morgan Kaufmann
  Publishers Inc.}, \bibinfo{address}{San Francisco, CA, USA},
  \bibinfo{pages}{1606–1611}.
\newblock
\urldef\tempurl%
\url{https://www.ijcai.org/Proceedings/07/Papers/259.pdf}
\showURL{%
\tempurl}


\bibitem[Gao et~al\mbox{.}(2021)]%
        {gao-etal-2021-simcse}
\bibfield{author}{\bibinfo{person}{Tianyu Gao}, \bibinfo{person}{Xingcheng
  Yao}, {and} \bibinfo{person}{Danqi Chen}.} \bibinfo{year}{2021}\natexlab{}.
\newblock \showarticletitle{{S}im{CSE}: Simple Contrastive Learning of Sentence
  Embeddings}. In \bibinfo{booktitle}{\emph{Proceedings of the 2021 Conference
  on Empirical Methods in Natural Language Processing}}.
  \bibinfo{publisher}{Association for Computational Linguistics},
  \bibinfo{address}{Online and Punta Cana, Dominican Republic},
  \bibinfo{pages}{6894--6910}.
\newblock
\urldef\tempurl%
\url{https://doi.org/10.18653/v1/2021.emnlp-main.552}
\showDOI{\tempurl}


\bibitem[Haj-Yahia et~al\mbox{.}(2019)]%
        {haj-yahia-etal-2019-towards}
\bibfield{author}{\bibinfo{person}{Zied Haj-Yahia}, \bibinfo{person}{Adrien
  Sieg}, {and} \bibinfo{person}{L{\'e}a~A. Deleris}.}
  \bibinfo{year}{2019}\natexlab{}.
\newblock \showarticletitle{Towards Unsupervised Text Classification Leveraging
  Experts and Word Embeddings}. In \bibinfo{booktitle}{\emph{Proceedings of the
  57th Annual Meeting of the Association for Computational Linguistics}}.
  \bibinfo{publisher}{Association for Computational Linguistics},
  \bibinfo{address}{Florence, Italy}, \bibinfo{pages}{371--379}.
\newblock
\urldef\tempurl%
\url{https://doi.org/10.18653/v1/P19-1036}
\showDOI{\tempurl}


\bibitem[Halder et~al\mbox{.}(2020)]%
        {halder-etal-2020-task}
\bibfield{author}{\bibinfo{person}{Kishaloy Halder}, \bibinfo{person}{Alan
  Akbik}, \bibinfo{person}{Josip Krapac}, {and} \bibinfo{person}{Roland
  Vollgraf}.} \bibinfo{year}{2020}\natexlab{}.
\newblock \showarticletitle{Task-Aware Representation of Sentences for Generic
  Text Classification}. In \bibinfo{booktitle}{\emph{Proceedings of the 28th
  International Conference on Computational Linguistics}}.
  \bibinfo{publisher}{International Committee on Computational Linguistics},
  \bibinfo{address}{Barcelona, Spain (Online)}, \bibinfo{pages}{3202--3213}.
\newblock
\urldef\tempurl%
\url{https://doi.org/10.18653/v1/2020.coling-main.285}
\showDOI{\tempurl}


\bibitem[He et~al\mbox{.}(2020)]%
        {DBLP:journals/corr/abs-2006-03654}
\bibfield{author}{\bibinfo{person}{Pengcheng He}, \bibinfo{person}{Xiaodong
  Liu}, \bibinfo{person}{Jianfeng Gao}, {and} \bibinfo{person}{Weizhu Chen}.}
  \bibinfo{year}{2020}\natexlab{}.
\newblock \showarticletitle{DeBERTa: Decoding-enhanced {BERT} with Disentangled
  Attention}.
\newblock \bibinfo{journal}{\emph{CoRR}}  \bibinfo{volume}{abs/2006.03654}
  (\bibinfo{year}{2020}).
\newblock
\showeprint[arXiv]{2006.03654}
\urldef\tempurl%
\url{https://arxiv.org/abs/2006.03654}
\showURL{%
\tempurl}


\bibitem[Kadhim(2019)]%
        {Kadhim2019}
\bibfield{author}{\bibinfo{person}{Ammar~Ismael Kadhim}.}
  \bibinfo{year}{2019}\natexlab{}.
\newblock \showarticletitle{Survey on supervised machine learning techniques
  for automatic text classification}.
\newblock \bibinfo{journal}{\emph{Artificial Intelligence Review}}
  \bibinfo{volume}{52}, \bibinfo{number}{1} (\bibinfo{date}{Jan.}
  \bibinfo{year}{2019}), \bibinfo{pages}{273--292}.
\newblock
\urldef\tempurl%
\url{https://doi.org/10.1007/s10462-018-09677-1}
\showDOI{\tempurl}


\bibitem[Kowsari et~al\mbox{.}(2019)]%
        {info10040150}
\bibfield{author}{\bibinfo{person}{Kamran Kowsari}, \bibinfo{person}{Kiana
  Jafari~Meimandi}, \bibinfo{person}{Mojtaba Heidarysafa},
  \bibinfo{person}{Sanjana Mendu}, \bibinfo{person}{Laura Barnes}, {and}
  \bibinfo{person}{Donald Brown}.} \bibinfo{year}{2019}\natexlab{}.
\newblock \showarticletitle{Text Classification Algorithms: A Survey}.
\newblock \bibinfo{journal}{\emph{Information}} \bibinfo{volume}{10},
  \bibinfo{number}{4} (\bibinfo{year}{2019}).
\newblock
\showISSN{2078-2489}
\urldef\tempurl%
\url{https://doi.org/10.3390/info10040150}
\showDOI{\tempurl}


\bibitem[Lang(1995)]%
        {Lang95}
\bibfield{author}{\bibinfo{person}{Ken Lang}.} \bibinfo{year}{1995}\natexlab{}.
\newblock \showarticletitle{Newsweeder: Learning to filter netnews}. In
  \bibinfo{booktitle}{\emph{Proceedings of the Twelfth International Conference
  on Machine Learning}}. \bibinfo{pages}{331--339}.
\newblock
\urldef\tempurl%
\url{https://citeseerx.ist.psu.edu/viewdoc/summary?doi=10.1.1.22.6286}
\showURL{%
\tempurl}


\bibitem[Le and Mikolov(2014)]%
        {pmlr-v32-le14}
\bibfield{author}{\bibinfo{person}{Quoc Le} {and} \bibinfo{person}{Tomas
  Mikolov}.} \bibinfo{year}{2014}\natexlab{}.
\newblock \showarticletitle{Distributed Representations of Sentences and
  Documents}. In \bibinfo{booktitle}{\emph{Proceedings of the 31st
  International Conference on Machine Learning}}
  \emph{(\bibinfo{series}{Proceedings of Machine Learning Research},
  Vol.~\bibinfo{volume}{32})}, \bibfield{editor}{\bibinfo{person}{Eric~P. Xing}
  {and} \bibinfo{person}{Tony Jebara}} (Eds.). \bibinfo{publisher}{PMLR},
  \bibinfo{address}{Bejing, China}, \bibinfo{pages}{1188--1196}.
\newblock
\urldef\tempurl%
\url{https://proceedings.mlr.press/v32/le14.html}
\showURL{%
\tempurl}


\bibitem[Lewis et~al\mbox{.}(2020)]%
        {lewis-etal-2020-bart}
\bibfield{author}{\bibinfo{person}{Mike Lewis}, \bibinfo{person}{Yinhan Liu},
  \bibinfo{person}{Naman Goyal}, \bibinfo{person}{Marjan Ghazvininejad},
  \bibinfo{person}{Abdelrahman Mohamed}, \bibinfo{person}{Omer Levy},
  \bibinfo{person}{Veselin Stoyanov}, {and} \bibinfo{person}{Luke
  Zettlemoyer}.} \bibinfo{year}{2020}\natexlab{}.
\newblock \showarticletitle{{BART}: Denoising Sequence-to-Sequence Pre-training
  for Natural Language Generation, Translation, and Comprehension}. In
  \bibinfo{booktitle}{\emph{Proceedings of the 58th Annual Meeting of the
  Association for Computational Linguistics}}. \bibinfo{publisher}{Association
  for Computational Linguistics}, \bibinfo{address}{Online},
  \bibinfo{pages}{7871--7880}.
\newblock
\urldef\tempurl%
\url{https://doi.org/10.18653/v1/2020.acl-main.703}
\showDOI{\tempurl}


\bibitem[Li et~al\mbox{.}(2016)]%
        {li-etal-2016-joint}
\bibfield{author}{\bibinfo{person}{Yuezhang Li}, \bibinfo{person}{Ronghuo
  Zheng}, \bibinfo{person}{Tian Tian}, \bibinfo{person}{Zhiting Hu},
  \bibinfo{person}{Rahul Iyer}, {and} \bibinfo{person}{Katia Sycara}.}
  \bibinfo{year}{2016}\natexlab{}.
\newblock \showarticletitle{Joint Embedding of Hierarchical Categories and
  Entities for Concept Categorization and Dataless Classification}. In
  \bibinfo{booktitle}{\emph{Proceedings of {COLING} 2016, the 26th
  International Conference on Computational Linguistics: Technical Papers}}.
  \bibinfo{publisher}{The COLING 2016 Organizing Committee},
  \bibinfo{address}{Osaka, Japan}, \bibinfo{pages}{2678--2688}.
\newblock
\urldef\tempurl%
\url{https://aclanthology.org/C16-1252}
\showURL{%
\tempurl}


\bibitem[Liu et~al\mbox{.}(2021)]%
        {9411914}
\bibfield{author}{\bibinfo{person}{Tengfei Liu}, \bibinfo{person}{Yongli Hu},
  \bibinfo{person}{Junbin Gao}, \bibinfo{person}{Yanfeng Sun}, {and}
  \bibinfo{person}{Baocai Yin}.} \bibinfo{year}{2021}\natexlab{}.
\newblock \showarticletitle{Zero-Shot Text Classification with Semantically
  Extended Graph Convolutional Network}. In \bibinfo{booktitle}{\emph{2020 25th
  International Conference on Pattern Recognition (ICPR)}}.
  \bibinfo{pages}{8352--8359}.
\newblock
\urldef\tempurl%
\url{https://doi.org/10.1109/ICPR48806.2021.9411914}
\showDOI{\tempurl}


\bibitem[Mikolov et~al\mbox{.}(2013)]%
        {NIPS2013_9aa42b31}
\bibfield{author}{\bibinfo{person}{Tomas Mikolov}, \bibinfo{person}{Ilya
  Sutskever}, \bibinfo{person}{Kai Chen}, \bibinfo{person}{Greg~S Corrado},
  {and} \bibinfo{person}{Jeff Dean}.} \bibinfo{year}{2013}\natexlab{}.
\newblock \showarticletitle{Distributed Representations of Words and Phrases
  and their Compositionality}. In \bibinfo{booktitle}{\emph{Advances in Neural
  Information Processing Systems}}, \bibfield{editor}{\bibinfo{person}{C.~J.~C.
  Burges}, \bibinfo{person}{L.~Bottou}, \bibinfo{person}{M.~Welling},
  \bibinfo{person}{Z.~Ghahramani}, {and} \bibinfo{person}{K.~Q. Weinberger}}
  (Eds.), Vol.~\bibinfo{volume}{26}. \bibinfo{publisher}{Curran Associates,
  Inc.}
\newblock
\urldef\tempurl%
\url{https://proceedings.neurips.cc/paper/2013/file/9aa42b31882ec039965f3c4923ce901b-Paper.pdf}
\showURL{%
\tempurl}


\bibitem[Minaee et~al\mbox{.}(2021)]%
        {10.1145/3439726}
\bibfield{author}{\bibinfo{person}{Shervin Minaee}, \bibinfo{person}{Nal
  Kalchbrenner}, \bibinfo{person}{Erik Cambria}, \bibinfo{person}{Narjes
  Nikzad}, \bibinfo{person}{Meysam Chenaghlu}, {and} \bibinfo{person}{Jianfeng
  Gao}.} \bibinfo{year}{2021}\natexlab{}.
\newblock \showarticletitle{Deep Learning--Based Text Classification: A
  Comprehensive Review}.
\newblock \bibinfo{journal}{\emph{ACM Comput. Surv.}} \bibinfo{volume}{54},
  \bibinfo{number}{3}, Article \bibinfo{articleno}{62} (\bibinfo{date}{apr}
  \bibinfo{year}{2021}), \bibinfo{numpages}{40}~pages.
\newblock
\showISSN{0360-0300}
\urldef\tempurl%
\url{https://doi.org/10.1145/3439726}
\showDOI{\tempurl}


\bibitem[Mirończuk and Protasiewicz(2018)]%
        {MIRONCZUK201836}
\bibfield{author}{\bibinfo{person}{Marcin~Michał Mirończuk} {and}
  \bibinfo{person}{Jarosław Protasiewicz}.} \bibinfo{year}{2018}\natexlab{}.
\newblock \showarticletitle{A recent overview of the state-of-the-art elements
  of text classification}.
\newblock \bibinfo{journal}{\emph{Expert Systems with Applications}}
  \bibinfo{volume}{106} (\bibinfo{year}{2018}), \bibinfo{pages}{36--54}.
\newblock
\showISSN{0957-4174}
\urldef\tempurl%
\url{https://doi.org/10.1016/j.eswa.2018.03.058}
\showDOI{\tempurl}


\bibitem[Nam et~al\mbox{.}(2016)]%
        {AAAI1612058}
\bibfield{author}{\bibinfo{person}{Jinseok Nam}, \bibinfo{person}{Eneldo~Loza
  Mencía}, {and} \bibinfo{person}{Johannes Fürnkranz}.}
  \bibinfo{year}{2016}\natexlab{}.
\newblock \showarticletitle{All-in Text: Learning Document, Label, and Word
  Representations Jointly}.
\newblock \bibinfo{journal}{\emph{AAAI Conference on Artificial Intelligence}}
  (\bibinfo{year}{2016}).
\newblock
\urldef\tempurl%
\url{https://www.aaai.org/ocs/index.php/AAAI/AAAI16/paper/view/12058}
\showURL{%
\tempurl}


\bibitem[Narayan et~al\mbox{.}(2020)]%
        {Narayan2020DensityPreservingDV}
\bibfield{author}{\bibinfo{person}{Ashwin Narayan}, \bibinfo{person}{Bonnie
  Berger}, {and} \bibinfo{person}{Hyunghoon Cho}.}
  \bibinfo{year}{2020}\natexlab{}.
\newblock \showarticletitle{Density-Preserving Data Visualization Unveils
  Dynamic Patterns of Single-Cell Transcriptomic Variability}.
\newblock \bibinfo{journal}{\emph{bioRxiv}} (\bibinfo{year}{2020}).
\newblock
\urldef\tempurl%
\url{https://doi.org/10.1101/2020.05.12.077776}
\showURL{%
\tempurl}


\bibitem[Parrish et~al\mbox{.}(2021)]%
        {parrish-etal-2021-putting-linguist}
\bibfield{author}{\bibinfo{person}{Alicia Parrish}, \bibinfo{person}{William
  Huang}, \bibinfo{person}{Omar Agha}, \bibinfo{person}{Soo-Hwan Lee},
  \bibinfo{person}{Nikita Nangia}, \bibinfo{person}{Alexia Warstadt},
  \bibinfo{person}{Karmanya Aggarwal}, \bibinfo{person}{Emily Allaway},
  \bibinfo{person}{Tal Linzen}, {and} \bibinfo{person}{Samuel~R. Bowman}.}
  \bibinfo{year}{2021}\natexlab{}.
\newblock \showarticletitle{Does Putting a Linguist in the Loop Improve {NLU}
  Data Collection?}. In \bibinfo{booktitle}{\emph{Findings of the Association
  for Computational Linguistics: EMNLP 2021}}. \bibinfo{publisher}{Association
  for Computational Linguistics}, \bibinfo{address}{Punta Cana, Dominican
  Republic}, \bibinfo{pages}{4886--4901}.
\newblock
\urldef\tempurl%
\url{https://doi.org/10.18653/v1/2021.findings-emnlp.421}
\showDOI{\tempurl}


\bibitem[Pushp and Srivastava(2017)]%
        {Pushp2017TrainOT}
\bibfield{author}{\bibinfo{person}{Pushpankar~Kumar Pushp} {and}
  \bibinfo{person}{Muktabh~Mayank Srivastava}.}
  \bibinfo{year}{2017}\natexlab{}.
\newblock \showarticletitle{Train Once, Test Anywhere: Zero-Shot Learning for
  Text Classification}.
\newblock \bibinfo{journal}{\emph{ArXiv}}  \bibinfo{volume}{abs/1712.05972}
  (\bibinfo{year}{2017}).
\newblock
\urldef\tempurl%
\url{https://arxiv.org/abs/1712.05972}
\showURL{%
\tempurl}


\bibitem[Radford et~al\mbox{.}(2019)]%
        {Radford2019LanguageMA}
\bibfield{author}{\bibinfo{person}{Alec Radford}, \bibinfo{person}{Jeff Wu},
  \bibinfo{person}{Rewon Child}, \bibinfo{person}{David Luan},
  \bibinfo{person}{Dario Amodei}, {and} \bibinfo{person}{Ilya Sutskever}.}
  \bibinfo{year}{2019}\natexlab{}.
\newblock \showarticletitle{Language Models are Unsupervised Multitask
  Learners}.
\newblock  (\bibinfo{year}{2019}).
\newblock
\urldef\tempurl%
\url{https://d4mucfpksywv.cloudfront.net/better-language-models/language-models.pdf}
\showURL{%
\tempurl}


\bibitem[Reimers and Gurevych(2019)]%
        {reimers-gurevych-2019-sentence}
\bibfield{author}{\bibinfo{person}{Nils Reimers} {and} \bibinfo{person}{Iryna
  Gurevych}.} \bibinfo{year}{2019}\natexlab{}.
\newblock \showarticletitle{Sentence-{BERT}: Sentence Embeddings using
  {S}iamese {BERT}-Networks}. In \bibinfo{booktitle}{\emph{Proceedings of the
  2019 Conference on Empirical Methods in Natural Language Processing and the
  9th International Joint Conference on Natural Language Processing
  (EMNLP-IJCNLP)}}. \bibinfo{publisher}{Association for Computational
  Linguistics}, \bibinfo{address}{Hong Kong, China},
  \bibinfo{pages}{3982--3992}.
\newblock
\urldef\tempurl%
\url{https://doi.org/10.18653/v1/D19-1410}
\showDOI{\tempurl}


\bibitem[Rios and Kavuluru(2018)]%
        {rios-kavuluru-2018-shot}
\bibfield{author}{\bibinfo{person}{Anthony Rios} {and}
  \bibinfo{person}{Ramakanth Kavuluru}.} \bibinfo{year}{2018}\natexlab{}.
\newblock \showarticletitle{Few-Shot and Zero-Shot Multi-Label Learning for
  Structured Label Spaces}. In \bibinfo{booktitle}{\emph{Proceedings of the
  2018 Conference on Empirical Methods in Natural Language Processing}}.
  \bibinfo{publisher}{Association for Computational Linguistics},
  \bibinfo{address}{Brussels, Belgium}, \bibinfo{pages}{3132--3142}.
\newblock
\urldef\tempurl%
\url{https://doi.org/10.18653/v1/D18-1352}
\showDOI{\tempurl}


\bibitem[Sanh et~al\mbox{.}(2019)]%
        {Sanh2019DistilBERTAD}
\bibfield{author}{\bibinfo{person}{Victor Sanh}, \bibinfo{person}{Lysandre
  Debut}, \bibinfo{person}{Julien Chaumond}, {and} \bibinfo{person}{Thomas
  Wolf}.} \bibinfo{year}{2019}\natexlab{}.
\newblock \showarticletitle{DistilBERT, a distilled version of BERT: smaller,
  faster, cheaper and lighter}.
\newblock \bibinfo{journal}{\emph{ArXiv}}  \bibinfo{volume}{abs/1910.01108}
  (\bibinfo{year}{2019}).
\newblock
\urldef\tempurl%
\url{https://arxiv.org/abs/1910.01108}
\showURL{%
\tempurl}


\bibitem[Sappadla et~al\mbox{.}(2016)]%
        {zero-shot-semsim}
\bibfield{author}{\bibinfo{person}{Prateek~Veeranna Sappadla},
  \bibinfo{person}{Jinseok Nam}, \bibinfo{person}{Eneldo~Loza Mencia}, {and}
  \bibinfo{person}{Johannes F\"urnkranz}.} \bibinfo{year}{2016}\natexlab{}.
\newblock \showarticletitle{Using semantic similarity for multi-label zero-shot
  classification of text documents}. In \bibinfo{booktitle}{\emph{Proc.
  European Sympo- sium on Artificial Neural Networks, Computational
  Intelligence and Machine Learning}}.
\newblock
\urldef\tempurl%
\url{https://www.esann.org/sites/default/files/proceedings/legacy/es2016-174.pdf}
\showURL{%
\tempurl}


\bibitem[Schneider et~al\mbox{.}(2022)]%
        {schneider-etal-2022-decade}
\bibfield{author}{\bibinfo{person}{Phillip Schneider}, \bibinfo{person}{Tim
  Schopf}, \bibinfo{person}{Juraj Vladika}, \bibinfo{person}{Mikhail Galkin},
  \bibinfo{person}{Elena Simperl}, {and} \bibinfo{person}{Florian Matthes}.}
  \bibinfo{year}{2022}\natexlab{}.
\newblock \showarticletitle{A Decade of Knowledge Graphs in Natural Language
  Processing: A Survey}. In \bibinfo{booktitle}{\emph{Proceedings of the 2nd
  Conference of the Asia-Pacific Chapter of the Association for Computational
  Linguistics and the 12th International Joint Conference on Natural Language
  Processing}}. \bibinfo{publisher}{Association for Computational Linguistics},
  \bibinfo{address}{Online only}, \bibinfo{pages}{601--614}.
\newblock
\urldef\tempurl%
\url{https://aclanthology.org/2022.aacl-main.46}
\showURL{%
\tempurl}


\bibitem[Schopf et~al\mbox{.}(2021)]%
        {webist21}
\bibfield{author}{\bibinfo{person}{Tim Schopf}, \bibinfo{person}{Daniel Braun},
  {and} \bibinfo{person}{Florian Matthes}.} \bibinfo{year}{2021}\natexlab{}.
\newblock \showarticletitle{Lbl2Vec: An Embedding-based Approach for
  Unsupervised Document Retrieval on Predefined Topics}. In
  \bibinfo{booktitle}{\emph{Proceedings of the 17th International Conference on
  Web Information Systems and Technologies - WEBIST,}}. INSTICC,
  \bibinfo{publisher}{SciTePress}, \bibinfo{pages}{124--132}.
\newblock
\showISBNx{978-989-758-536-4}
\showISSN{2184-3252}
\urldef\tempurl%
\url{https://doi.org/10.5220/0010710300003058}
\showDOI{\tempurl}


\bibitem[Schopf et~al\mbox{.}(2023)]%
        {schopf_lbl2vec_23}
\bibfield{author}{\bibinfo{person}{Tim Schopf}, \bibinfo{person}{Daniel Braun},
  {and} \bibinfo{person}{Florian Matthes}.} \bibinfo{year}{2023}\natexlab{}.
\newblock \showarticletitle{Semantic Label Representations with Lbl2Vec: A
  Similarity-Based Approach for Unsupervised Text Classification}. In
  \bibinfo{booktitle}{\emph{Web Information Systems and Technologies}},
  \bibfield{editor}{\bibinfo{person}{Massimo Marchiori},
  \bibinfo{person}{Francisco~Jos{\'e} Dom{\'i}nguez~Mayo}, {and}
  \bibinfo{person}{Joaquim Filipe}} (Eds.). \bibinfo{publisher}{Springer
  International Publishing}, \bibinfo{address}{Cham}, \bibinfo{pages}{59--73}.
\newblock
\showISBNx{978-3-031-24197-0}


\bibitem[Schopf et~al\mbox{.}(2022a)]%
        {schopf_etal_kdir22}
\bibfield{author}{\bibinfo{person}{Tim Schopf}, \bibinfo{person}{Simon Klimek},
  {and} \bibinfo{person}{Florian Matthes}.} \bibinfo{year}{2022}\natexlab{a}.
\newblock \showarticletitle{PatternRank: Leveraging Pretrained Language Models
  and Part of Speech for Unsupervised Keyphrase Extraction}. In
  \bibinfo{booktitle}{\emph{Proceedings of the 14th International Joint
  Conference on Knowledge Discovery, Knowledge Engineering and Knowledge
  Management - KDIR}}. INSTICC, \bibinfo{publisher}{SciTePress},
  \bibinfo{pages}{243--248}.
\newblock
\showISBNx{978-989-758-614-9}
\showISSN{2184-3228}
\urldef\tempurl%
\url{https://doi.org/10.5220/0011546600003335}
\showDOI{\tempurl}


\bibitem[Schopf et~al\mbox{.}(2022b)]%
        {10.1145/3535782.3535835}
\bibfield{author}{\bibinfo{person}{Tim Schopf}, \bibinfo{person}{Peter
  Weinberger}, \bibinfo{person}{Thomas Kinkeldei}, {and}
  \bibinfo{person}{Florian Matthes}.} \bibinfo{year}{2022}\natexlab{b}.
\newblock \showarticletitle{Towards Bilingual Word Embedding Models for
  Engineering: Evaluating Semantic Linking Capabilities of Engineering-Specific
  Word Embeddings Across Languages}. In \bibinfo{booktitle}{\emph{2022 4th
  International Conference on Management Science and Industrial Engineering
  (MSIE)}} (Chiang Mai, Thailand) \emph{(\bibinfo{series}{MSIE 2022})}.
  \bibinfo{publisher}{Association for Computing Machinery},
  \bibinfo{address}{New York, NY, USA}, \bibinfo{pages}{407–413}.
\newblock
\showISBNx{9781450395816}
\urldef\tempurl%
\url{https://doi.org/10.1145/3535782.3535835}
\showDOI{\tempurl}


\bibitem[Song and Roth(2014)]%
        {Song_Roth_2014}
\bibfield{author}{\bibinfo{person}{Yangqiu Song} {and} \bibinfo{person}{Dan
  Roth}.} \bibinfo{year}{2014}\natexlab{}.
\newblock \showarticletitle{On Dataless Hierarchical Text Classification}.
\newblock \bibinfo{journal}{\emph{Proceedings of the AAAI Conference on
  Artificial Intelligence}} \bibinfo{volume}{28}, \bibinfo{number}{1}
  (\bibinfo{date}{Jun.} \bibinfo{year}{2014}).
\newblock
\urldef\tempurl%
\url{https://ojs.aaai.org/index.php/AAAI/article/view/8938}
\showURL{%
\tempurl}


\bibitem[Song et~al\mbox{.}(2016)]%
        {10.5555/3060832.3061027}
\bibfield{author}{\bibinfo{person}{Yangqiu Song}, \bibinfo{person}{Shyam
  Upadhyay}, \bibinfo{person}{Haoruo Peng}, {and} \bibinfo{person}{Dan Roth}.}
  \bibinfo{year}{2016}\natexlab{}.
\newblock \showarticletitle{Cross-Lingual Dataless Classification for Many
  Languages}. In \bibinfo{booktitle}{\emph{Proceedings of the Twenty-Fifth
  International Joint Conference on Artificial Intelligence}} (New York, New
  York, USA) \emph{(\bibinfo{series}{IJCAI'16})}. \bibinfo{publisher}{AAAI
  Press}, \bibinfo{pages}{2901–2907}.
\newblock
\showISBNx{9781577357704}
\urldef\tempurl%
\url{https://www.ijcai.org/Proceedings/16/Papers/412.pdf}
\showURL{%
\tempurl}


\bibitem[Stammbach and Ash(2021)]%
        {Stammbach2021DocSCANUT}
\bibfield{author}{\bibinfo{person}{Dominik Stammbach} {and}
  \bibinfo{person}{Elliott Ash}.} \bibinfo{year}{2021}\natexlab{}.
\newblock \showarticletitle{DocSCAN: Unsupervised Text Classification via
  Learning from Neighbors}.
\newblock \bibinfo{journal}{\emph{ArXiv}}  \bibinfo{volume}{abs/2105.04024}
  (\bibinfo{year}{2021}).
\newblock
\urldef\tempurl%
\url{https://arxiv.org/abs/2105.04024}
\showURL{%
\tempurl}


\bibitem[Thorne et~al\mbox{.}(2018)]%
        {thorne-etal-2018-fever}
\bibfield{author}{\bibinfo{person}{James Thorne}, \bibinfo{person}{Andreas
  Vlachos}, \bibinfo{person}{Christos Christodoulopoulos}, {and}
  \bibinfo{person}{Arpit Mittal}.} \bibinfo{year}{2018}\natexlab{}.
\newblock \showarticletitle{{FEVER}: a Large-scale Dataset for Fact Extraction
  and {VER}ification}. In \bibinfo{booktitle}{\emph{Proceedings of the 2018
  Conference of the North {A}merican Chapter of the Association for
  Computational Linguistics: Human Language Technologies, Volume 1 (Long
  Papers)}}. \bibinfo{publisher}{Association for Computational Linguistics},
  \bibinfo{address}{New Orleans, Louisiana}, \bibinfo{pages}{809--819}.
\newblock
\urldef\tempurl%
\url{https://doi.org/10.18653/v1/N18-1074}
\showDOI{\tempurl}


\bibitem[Wang et~al\mbox{.}(2019)]%
        {10.1145/3293318}
\bibfield{author}{\bibinfo{person}{Wei Wang}, \bibinfo{person}{Vincent~W.
  Zheng}, \bibinfo{person}{Han Yu}, {and} \bibinfo{person}{Chunyan Miao}.}
  \bibinfo{year}{2019}\natexlab{}.
\newblock \showarticletitle{A Survey of Zero-Shot Learning: Settings, Methods,
  and Applications}.
\newblock \bibinfo{journal}{\emph{ACM Trans. Intell. Syst. Technol.}}
  \bibinfo{volume}{10}, \bibinfo{number}{2}, Article \bibinfo{articleno}{13}
  (\bibinfo{date}{jan} \bibinfo{year}{2019}), \bibinfo{numpages}{37}~pages.
\newblock
\showISSN{2157-6904}
\urldef\tempurl%
\url{https://doi.org/10.1145/3293318}
\showDOI{\tempurl}


\bibitem[Williams et~al\mbox{.}(2018)]%
        {williams-etal-2018-broad}
\bibfield{author}{\bibinfo{person}{Adina Williams}, \bibinfo{person}{Nikita
  Nangia}, {and} \bibinfo{person}{Samuel Bowman}.}
  \bibinfo{year}{2018}\natexlab{}.
\newblock \showarticletitle{A Broad-Coverage Challenge Corpus for Sentence
  Understanding through Inference}. In \bibinfo{booktitle}{\emph{Proceedings of
  the 2018 Conference of the North {A}merican Chapter of the Association for
  Computational Linguistics: Human Language Technologies, Volume 1 (Long
  Papers)}}. \bibinfo{publisher}{Association for Computational Linguistics},
  \bibinfo{address}{New Orleans, Louisiana}, \bibinfo{pages}{1112--1122}.
\newblock
\urldef\tempurl%
\url{https://doi.org/10.18653/v1/N18-1101}
\showDOI{\tempurl}


\bibitem[Ye et~al\mbox{.}(2020)]%
        {ye-etal-2020-zero}
\bibfield{author}{\bibinfo{person}{Zhiquan Ye}, \bibinfo{person}{Yuxia Geng},
  \bibinfo{person}{Jiaoyan Chen}, \bibinfo{person}{Jingmin Chen},
  \bibinfo{person}{Xiaoxiao Xu}, \bibinfo{person}{SuHang Zheng},
  \bibinfo{person}{Feng Wang}, \bibinfo{person}{Jun Zhang}, {and}
  \bibinfo{person}{Huajun Chen}.} \bibinfo{year}{2020}\natexlab{}.
\newblock \showarticletitle{Zero-shot Text Classification via Reinforced
  Self-training}. In \bibinfo{booktitle}{\emph{Proceedings of the 58th Annual
  Meeting of the Association for Computational Linguistics}}.
  \bibinfo{publisher}{Association for Computational Linguistics},
  \bibinfo{address}{Online}, \bibinfo{pages}{3014--3024}.
\newblock
\urldef\tempurl%
\url{https://doi.org/10.18653/v1/2020.acl-main.272}
\showDOI{\tempurl}


\bibitem[Yin et~al\mbox{.}(2019)]%
        {yin-etal-2019-benchmarking}
\bibfield{author}{\bibinfo{person}{Wenpeng Yin}, \bibinfo{person}{Jamaal Hay},
  {and} \bibinfo{person}{Dan Roth}.} \bibinfo{year}{2019}\natexlab{}.
\newblock \showarticletitle{Benchmarking Zero-shot Text Classification:
  Datasets, Evaluation and Entailment Approach}. In
  \bibinfo{booktitle}{\emph{Proceedings of the 2019 Conference on Empirical
  Methods in Natural Language Processing and the 9th International Joint
  Conference on Natural Language Processing (EMNLP-IJCNLP)}}.
  \bibinfo{publisher}{Association for Computational Linguistics},
  \bibinfo{address}{Hong Kong, China}, \bibinfo{pages}{3914--3923}.
\newblock
\urldef\tempurl%
\url{https://doi.org/10.18653/v1/D19-1404}
\showDOI{\tempurl}


\bibitem[Yin et~al\mbox{.}(2021)]%
        {yin-etal-2021-docnli}
\bibfield{author}{\bibinfo{person}{Wenpeng Yin}, \bibinfo{person}{Dragomir
  Radev}, {and} \bibinfo{person}{Caiming Xiong}.}
  \bibinfo{year}{2021}\natexlab{}.
\newblock \showarticletitle{{D}oc{NLI}: A Large-scale Dataset for
  Document-level Natural Language Inference}. In
  \bibinfo{booktitle}{\emph{Findings of the Association for Computational
  Linguistics: ACL-IJCNLP 2021}}. \bibinfo{publisher}{Association for
  Computational Linguistics}, \bibinfo{address}{Online},
  \bibinfo{pages}{4913--4922}.
\newblock
\urldef\tempurl%
\url{https://doi.org/10.18653/v1/2021.findings-acl.435}
\showDOI{\tempurl}


\bibitem[Zhang et~al\mbox{.}(2019)]%
        {zhang-etal-2019-integrating}
\bibfield{author}{\bibinfo{person}{Jingqing Zhang}, \bibinfo{person}{Piyawat
  Lertvittayakumjorn}, {and} \bibinfo{person}{Yike Guo}.}
  \bibinfo{year}{2019}\natexlab{}.
\newblock \showarticletitle{Integrating Semantic Knowledge to Tackle Zero-shot
  Text Classification}. In \bibinfo{booktitle}{\emph{Proceedings of the 2019
  Conference of the North {A}merican Chapter of the Association for
  Computational Linguistics: Human Language Technologies, Volume 1 (Long and
  Short Papers)}}. \bibinfo{publisher}{Association for Computational
  Linguistics}, \bibinfo{address}{Minneapolis, Minnesota},
  \bibinfo{pages}{1031--1040}.
\newblock
\urldef\tempurl%
\url{https://doi.org/10.18653/v1/N19-1108}
\showDOI{\tempurl}


\bibitem[Zhang et~al\mbox{.}(2015)]%
        {10.5555/2969239.2969312}
\bibfield{author}{\bibinfo{person}{Xiang Zhang}, \bibinfo{person}{Junbo Zhao},
  {and} \bibinfo{person}{Yann LeCun}.} \bibinfo{year}{2015}\natexlab{}.
\newblock \showarticletitle{Character-Level Convolutional Networks for Text
  Classification}. In \bibinfo{booktitle}{\emph{Proceedings of the 28th
  International Conference on Neural Information Processing Systems - Volume
  1}} (Montreal, Canada) \emph{(\bibinfo{series}{NIPS'15})}.
  \bibinfo{publisher}{MIT Press}, \bibinfo{address}{Cambridge, MA, USA},
  \bibinfo{pages}{649–657}.
\newblock
\urldef\tempurl%
\url{https://proceedings.neurips.cc/paper/2015/file/250cf8b51c773f3f8dc8b4be867a9a02-Paper.pdf}
\showURL{%
\tempurl}


\end{thebibliography}

\newpage
\appendix
\section{Appendix}
\label{sec:appendix}

\subsection{20Newsgroups Class Summary}\label{sec:appendix_20news}

\begin{table}[!h]
    \centering
    \scalebox{0.95}{
    \begin{tabular}{|l|l|}
    \hline
    \multicolumn{1}{|c|}{\textbf{Class Name}} & \multicolumn{1}{c|}{\textbf{Label Keywords}}                                  \\ \hline
    alt.atheism                               & atheism                                                                       \\ \hline
    comp.graphics                             & computer, graphics                                                            \\ \hline
    comp.os.ms-windows.misc                   & \begin{tabular}[c]{@{}l@{}}computer, os,\\ microsoft, windows\end{tabular}    \\ \hline
    comp.sys.ibm.pc.hardware                  & \begin{tabular}[c]{@{}l@{}}computer, system,\\ ibm, pc, hardware\end{tabular} \\ \hline
    comp.sys.mac.hardware                     & \begin{tabular}[c]{@{}l@{}}computer, system,\\ mac, hardware\end{tabular}     \\ \hline
    comp.windows.x                            & computer, windows                                                             \\ \hline
    misc.forsale                              & forsale                                                                       \\ \hline
    rec.autos                                 & cars                                                                          \\ \hline
    rec.motorcycles                           & motorcycles                                                                   \\ \hline
    rec.sport.baseball                        & sport, baseball                                                               \\ \hline
    rec.sport.hockey                          & sport, hockey                                                                 \\ \hline
    sci.crypt                                 & encryption                                                                    \\ \hline
    sci.electronics                           & electronics                                                                   \\ \hline
    sci.med                                   & medical                                                                       \\ \hline
    sci.space                                 & space                                                                         \\ \hline
    soc.religion.christian                    & religion, christianity                                                        \\ \hline
    talk.politics.guns                        & politics, guns                                                                \\ \hline
    talk.politics.mideast                     & politics, arab                                                                \\ \hline
    talk.politics.misc                        & politics                                                                      \\ \hline
    talk.religion.misc                        & religion                                                                      \\ \hline
    \end{tabular}}
    \caption{\label{tab:20news}20Newsgroups class names and inferred label keywords.}
\end{table}

\subsection{AG's Corpus Class Summary}\label{sec:appendix_ag}

\begin{table}[!h]
    \centering
    \scalebox{1}{
    \begin{tabular}{|l|l|}
    \hline
    \multicolumn{1}{|c|}{\textbf{Class Name}} & \multicolumn{1}{c|}{\textbf{Label Keywords}} \\ \hline
    World                                     & government                                   \\ \hline
    Sports                                    & sports                                       \\ \hline
    Business                                  & business                                     \\ \hline
    Science/Technology                        & science, technology                          \\ \hline
    \end{tabular}}
    \caption{\label{tab:agnews}AG's Corpus class names and inferred label keywords.}
\end{table}

\newpage

\subsection{Yahoo! Answers Class Summary}\label{sec:appendix_yahoo}

\begin{table}[!h]
    \centering
    \scalebox{1}{
    \begin{tabular}{|l|l|}
    \hline
    \multicolumn{1}{|c|}{\textbf{Class Name}} & \multicolumn{1}{c|}{\textbf{Label Keywords}} \\ \hline
    Society \& Culture                        & society, culture                             \\ \hline
    Science \& Mathematics                    & science, mathematics                         \\ \hline
    Health                                    & health                                       \\ \hline
    Education \& Reference                    & education, reference                         \\ \hline
    Computers \& Internet                     & computers, internet                          \\ \hline
    Sports                                    & sports                                       \\ \hline
    Business \& Finance                       & business, finance                            \\ \hline
    Entertainment \& Music                    & entertainment, music                         \\ \hline
    Family \& Relationships                   & family, relationships                        \\ \hline
    Politics \& Government                    & politics, government                         \\ \hline
    \end{tabular}}
    \caption{\label{tab:yahoo}Yahoo! Answers class names and inferred label keywords.}
\end{table}

\subsection{Medical Abstracts Class Summary}\label{sec:appendix_medical}

\begin{table}[!h]
    \centering
    \scalebox{1}{
    \begin{tabular}{|l|l|}
    \hline
    \multicolumn{1}{|c|}{\textbf{Class Name}} & \multicolumn{1}{c|}{\textbf{Label Keywords}} \\ \hline
    Neoplasms                        & neoplasms                                    \\ \hline
    \begin{tabular}[c]{@{}l@{}}Digestive system\\ diseases\end{tabular}        &   \begin{tabular}[c]{@{}l@{}}intestine, system, \\ diseases\end{tabular}                \\ \hline
    \begin{tabular}[c]{@{}l@{}}Nervous system\\ diseases\end{tabular}          & \begin{tabular}[c]{@{}l@{}}nervous, system, \\ diseases\end{tabular}                    \\ \hline
    \begin{tabular}[c]{@{}l@{}}Cardiovascular\\ diseases\end{tabular}          & \begin{tabular}[c]{@{}l@{}}cardiovascular,\\ diseases\end{tabular}                     \\ \hline
    \begin{tabular}[c]{@{}l@{}}General pathological\\ conditions\end{tabular}  & \begin{tabular}[c]{@{}l@{}}general, pathological,\\ conditions\end{tabular}            \\ \hline
    \end{tabular}}
    \caption{\label{tab:medical}Medical Abstracts class names and inferred label keywords.}
\end{table}
\clearpage
\subsection{DensMAP Dataset Visualizations}\label{sec:appendix_umap}

\centering
\begin{minipage}[t]{2\linewidth}\centering
    \includegraphics[width=\linewidth]{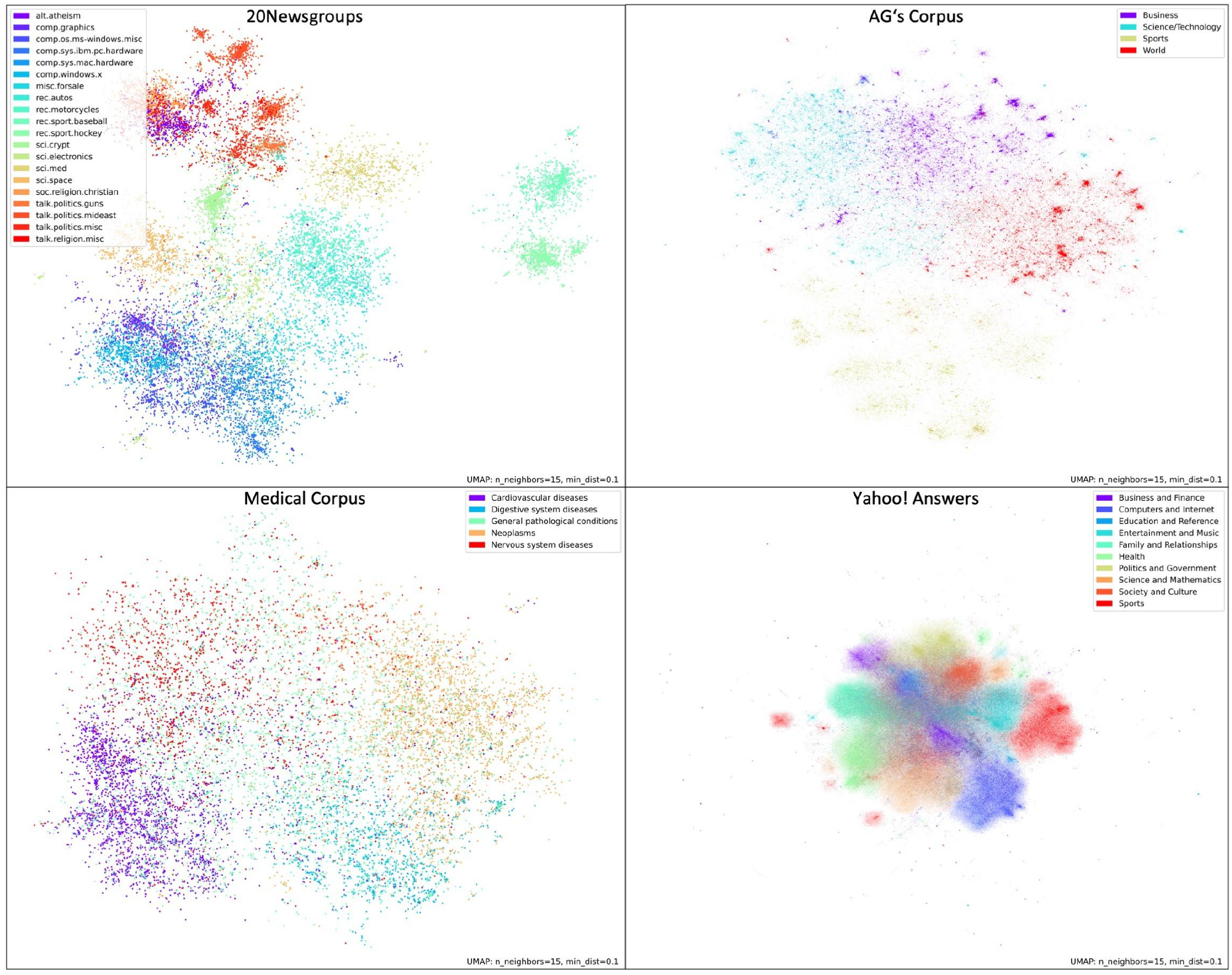}
    \captionsetup{width=\linewidth}
    \hspace{2em} \captionof{figure}{DensMAP visualizations of the document representations for each dataset described in Section \ref{sec:datasets}. The document representations were created by applying the average paragraph embedding strategy described in Section \ref{sec:baselines} using SBERT (\textit{all-mpnet-base-v2}).\label{fig:densmap}}
\end{minipage}

\end{document}